\documentclass{article}

\usepackage{microtype}
\usepackage{graphicx}
\usepackage{subcaption}
\usepackage{booktabs} 

\usepackage{hyperref}

\usepackage[accepted]{icml2026}

\usepackage{amsmath}
\usepackage{amssymb}
\usepackage{mathtools}
\usepackage{amsthm}

\usepackage[capitalize,noabbrev]{cleveref}

\theoremstyle{plain}

\theoremstyle{definition}

\theoremstyle{remark}

\usepackage{tcolorbox}
\usepackage{multirow}
\usepackage[dvipsnames]{xcolor} 

\newcommand{\gain}[1]{{\color{ForestGreen}\small{(+#1)}}}
\newcommand{\loss}[1]{{\color{BrickRed}\small{(#1)}}}    
                               
\usepackage[textsize=tiny]{todonotes}

\icmltitlerunning{TRE: Encouraging Exploration in the Trust Region}
\makeatletter
\renewcommand{\ICML@appearing}{Preprint.}
\makeatother
\begin{document}

\twocolumn[
  \icmltitle{TRE: Encouraging Exploration in the Trust Region}



  \icmlsetsymbol{equal}{*}

  \begin{icmlauthorlist}
    \icmlauthor{Chao Huang}{iie,ucas}
    \icmlauthor{Yujing Lu}{baidu}
    \icmlauthor{Quangang Li}{iie,ucas}
    \icmlauthor{Shenghe Wang}{baidu}
    \icmlauthor{Yan Wang}{baidu}
    \icmlauthor{Yueyang Zhang}{baidu}
    \icmlauthor{Long Xia}{baidu}
    \icmlauthor{Jiashu Zhao}{wlu}
    \icmlauthor{Zhiyuan Sun}{baidu}
    \icmlauthor{Daiting Shi}{baidu}
    \icmlauthor{Tingwen Liu}{iie,ucas}
  \end{icmlauthorlist}

    \icmlaffiliation{iie}{Institute of Information Engineering, Chinese Academy of Sciences}
    \icmlaffiliation{ucas}{School of Cyber Security, University of Chinese Academy of Sciences}
    \icmlaffiliation{baidu}{Baidu Inc.}
    \icmlaffiliation{wlu}{Wilfrid Laurier University}

\begin{center}
    \small \ttfamily
    \{huangchao, liquangang, liutingwen\}@iie.ac.cn, jzhao@wlu.ca \\
    \{luyujing, wangshenghe01, wangyan78, zhangyueyang, \\ 
    xialong01, sunzhiyuan01, shidaiting01\}@baidu.com
\end{center}
  \icmlcorrespondingauthor{Chao Huang}{huangchao@iie.ac.cn}
  


  \vskip 0.3in
]



\printAffiliationsAndNotice{}  

\begin{abstract}
Entropy regularization is a standard technique in reinforcement learning (RL) to enhance exploration, yet it yields negligible effects or even degrades performance in Large Language Models (LLMs). We attribute this failure to the \textbf{cumulative tail risk} inherent to LLMs with massive vocabularies and long generation horizons. In such environments, standard global entropy maximization indiscriminately dilutes probability mass into the vast tail of invalid tokens rather than focusing on plausible candidates, thereby disrupting coherent reasoning. To address this, we propose \textbf{Trust Region Entropy (TRE)}, a method that encourages exploration strictly within the model's trust region. Extensive experiments across mathematical reasoning (MATH), combinatorial search (Countdown), and preference alignment (HH) tasks demonstrate that TRE consistently outperforms vanilla PPO, standard entropy regularization, and other exploration baselines. Our code is available at \href{https://github.com/WhyChaos/TRE-Encouraging-Exploration-in-the-Trust-Region}{github.com/WhyChaos/TRE}.

\end{abstract}

\section{Introduction}
\label{sec:introduction}

Reinforcement Learning (RL) has emerged as a cornerstone paradigm for aligning Large Language Models (LLMs) with human intent and enhancing their capabilities in complex reasoning tasks \citep{guo2025deepseek,ouyang2022training}. Whether fine-tuning for helpfulness or optimizing for correct mathematical derivation, the central challenge remains the efficient exploration of the policy space. In classical RL domains—such as robotics or grid-world navigation—this exploration-exploitation trade-off is typically managed via \textbf{entropy regularization}, which penalizes deterministic policies and encourages the agent to sample a diverse range of actions \citep{ashlag2025state,ahmed2019understanding,haarnoja2018soft}.

However, despite its ubiquity in standard control problems, the direct application of entropy regularization to LLMs has shown limited benefits and often degrades performance \citep{cui2025entropy}. We posit that this failure stems from the distinct interaction between the \textbf{sparsity of valid manifolds} and the \textbf{cumulative risk in long-horizon generation}. Unlike classical RL environments where random exploration often remains within a valid state space, the valid tokens in LLM generation constitute an infinitesimally sparse manifold within a massive vocabulary. Standard entropy regularization, by encouraging global uniformity, indiscriminately shifts probability mass from plausible candidates into the vast tail of syntactically or logically invalid tokens. While this local noise might be negligible in single-step decisions, LLM reasoning requires maintaining coherence over long sequences. As the generation horizon extends, this injected noise accumulates exponentially, rendering the maintenance of a valid Chain-of-Thought statistically impossible. Consequently, global entropy maximization functions less as structured exploration and more as a disruptor of long-chain reasoning.We provide a detailed empirical analysis of these failure modes and their correlation with generation length in Section~\ref{sec:method_failure}.

To address this dimensionality mismatch, we propose \textbf{Trust Region Entropy (TRE)}, a novel regularization framework designed specifically for the sparse nature of language generation. TRE operates on the intuition that valid exploration should be constrained to a trust region—the subset of tokens that the pre-trained model considers plausible. Rather than flattening the entire distribution, TRE maximizes entropy strictly within this high-confidence subset(see Section~\ref{sec:method}). 

Our contributions are summarized as follows:
\begin{itemize}
    \item We provide an empirical analysis revealing that the failure of standard entropy regularization is driven by \textbf{cumulative tail risk} in long-horizon generation. We demonstrate that while regularization may offer marginal benefits in short-horizon settings, it becomes increasingly detrimental as the generation budget extends, explaining why reasoning tasks are particularly vulnerable to global entropy maximization.
    \item We introduce \textbf{Trust Region Entropy (TRE)}, a method that encourages exploration strictly within a trust region via local entropy maximization.
    \item We demonstrate through extensive experiments on mathematical reasoning (MATH), combinatorial search (Countdown), and preference alignment (HH) that TRE consistently outperforms vanilla PPO, standard entropy regularization, and other exploration baselines.
\end{itemize}

\section{Related Work}
\label{sec:related_work}

\paragraph{RL for LLM Alignment}
Reinforcement Learning (RL) has become indispensable for aligning Large Language Models (LLMs) with complex human values and reasoning requirements. Following the standard Reinforcement Learning from Human Feedback (RLHF) pipeline \citep{ouyang2022training}, models initially trained via supervised fine-tuning are further optimized using algorithms such as Proximal Policy Optimization (PPO) \citep{schulman2017proximal} to maximize non-differentiable reward signals. This paradigm has proven effective across various domains, from improving helpfulness and safety \citep{bai2022training} to enhancing mathematical reasoning capabilities \citep{guo2025deepseek,yu2025dapo}. 

\paragraph{Entropy Regularization}
Entropy regularization is a cornerstone technique in modern RL, encouraging exploration via the entropy term. While highly effective in low-dimensional continuous control, naive entropy maximization proves problematic for LLMs due to massive vocabulary sizes~\cite{cui2025entropy}. To mitigate this, contemporaneous works have proposed selective constraint mechanisms. For instance, \citet{wang2025beyond} propose \textit{Forking-Tokens}, which restricts optimization to steps with high entropy to preserve exploratory potential. Similarly, \citet{cui2025entropy} introduces \textit{KL-Cov}, which identifies steps with high covariance between advantage estimates and log-probabilities, selectively imposing a strong KL penalty on these critical steps to stabilize training dynamics. Specific implementation details of these baselines are provided in Section~\ref{sec:setup}.

\paragraph{Trust Region}
The concept of a \textit{Trust Region} is foundational to stable optimization in reinforcement learning. Trust Region Policy Optimization (TRPO) \citep{schulman2015trust} constrains the policy update by enforcing a strict KL-divergence constraint on a surrogate objective, ensuring monotonic improvement while maintaining stability. This surrogate objective is designed to approximate the true objective while keeping the updates within a trust region defined by the KL-divergence. In contrast, PPO \citep{schulman2017proximal} simplifies this approach by introducing a clipped surrogate objective that penalizes large policy deviations, making it more tractable and efficient, while still achieving similar stability to TRPO.

In the context of language generation, inference strategies like top-$k$ or Nucleus Sampling (top-$p$) truncate the low-probability tail during decoding. As \citet{Holtzman2020The} note, this truncation strategy can be interpreted as determining the model's trust region. Our Trust Region Entropy (TRE) adapts this established trust region concept to the \textit{training-time exploration objective}.

\begin{figure*}[t]
    \centering
    \begin{minipage}{0.49\textwidth}
        \centering
        \includegraphics[width=\linewidth]{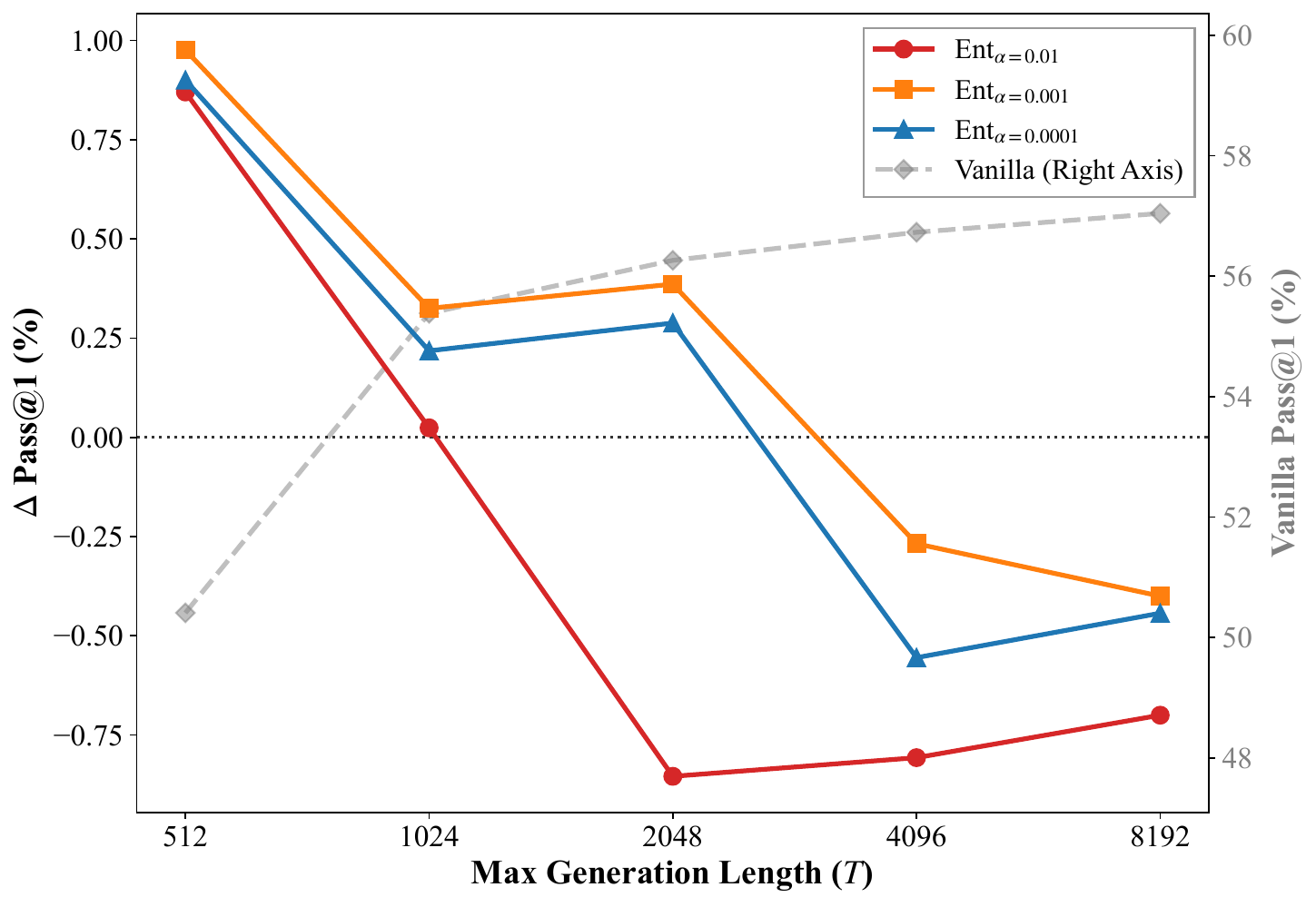}
        \centerline{\small \textbf{(a) MATH Task}}
    \end{minipage}
    \hfill
    \begin{minipage}{0.49\textwidth}
        \centering
        \includegraphics[width=\linewidth]{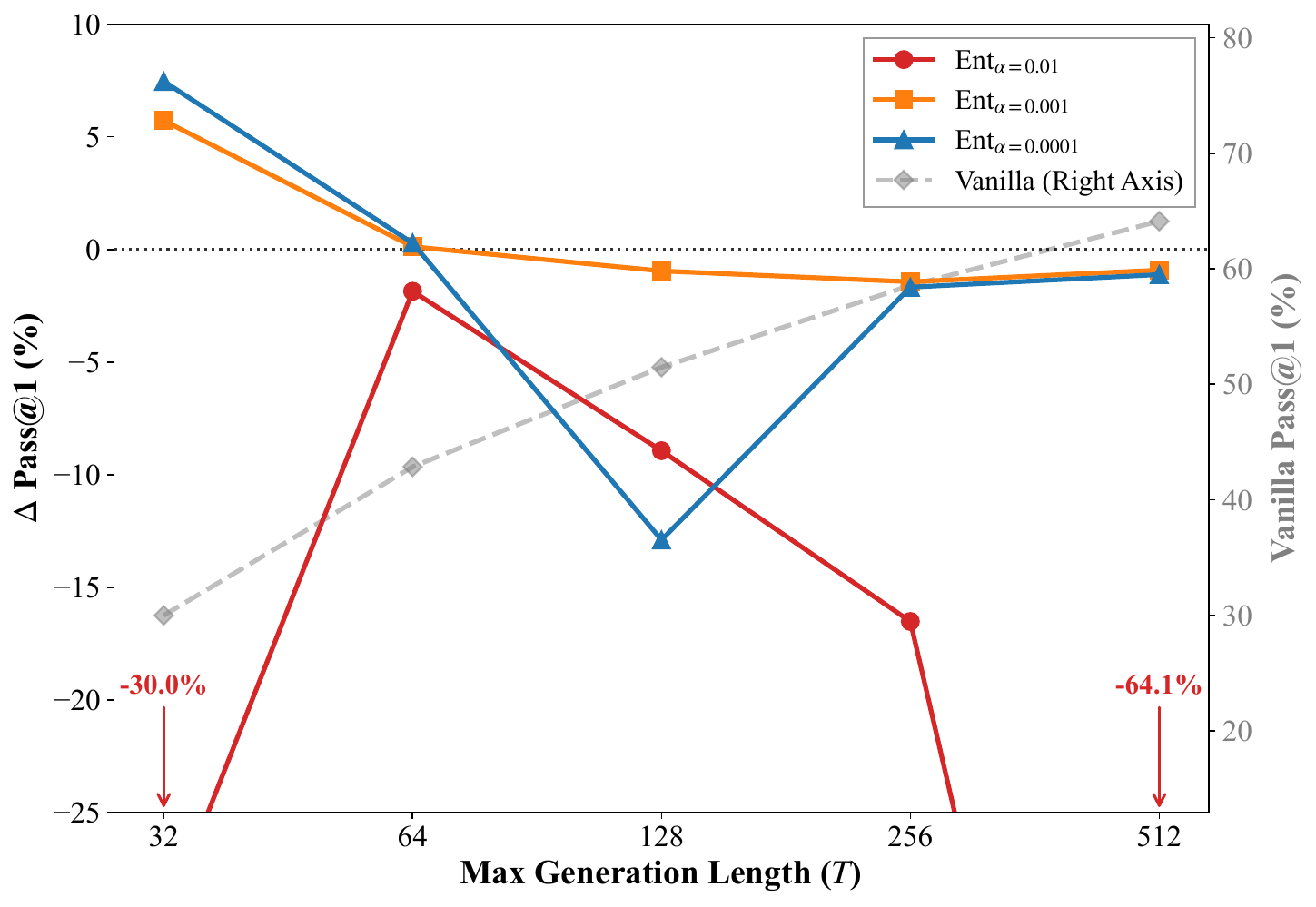}
        \centerline{\small \textbf{(b) Countdown Task}}
    \end{minipage}
    
    \vspace{-0.2cm}
    
    \caption{\textbf{Impact of Maximum Generation Length ($T$) on Entropy Regularization Efficacy.} 
    Experiments are conducted using Qwen2.5-1.5B-Instruct.
    The left $y$-axis displays the performance difference ($\Delta$) between entropy-regularized PPO and the vanilla baseline. The grey dashed line (corresponding to the right $y$-axis) depicts the absolute accuracy of the vanilla baseline. 
    Across both domains, mild entropy regularization helps in short-horizon settings but becomes detrimental as the horizon extends, showing its incompatibility with long-context reasoning.}
    \label{fig:entropy_length_comparison}
\end{figure*}


\section{Preliminaries}
\label{sec:preliminaries}

\paragraph{RL for LLMs}
We formulate the Large Language Model (LLM) generation process as a sequential decision-making problem. The LLM functions as a parameterized softmax policy $\pi_\theta$ over a discrete action space $\mathcal{A}$ (the vocabulary), with size $|\mathcal{A}|$.

At each time step $t$, given the current context $s_t$ (comprising the prompt $q$ and history $a_{<t}$), the underlying neural network $f_\theta$ maps the state to a logit vector $\mathbf{z}_t = f_\theta(s_t) \in \mathbb{R}^{|\mathcal{A}|}$. The policy distribution is defined by applying the softmax function to these logits:
\begin{equation}
\pi_\theta(a \mid s_t) = \frac{\exp(z_{t,a})}{\sum_{a' \in \mathcal{A}} \exp(z_{t,a'})},
\label{eq:softmax_policy}
\end{equation}
where $z_{t,a}$ denotes the logit value associated with token $a$.

For a complete response sequence $a = (a_1, \dots, a_T)$, the joint probability is factorized autoregressively:
\begin{equation}
\pi_\theta(a \mid q) = \prod_{t=1}^{T} \pi_\theta(a_t \mid s_t).
\label{eq:autoregressive_policy}
\end{equation}

We model this process as an episodic Markov Decision Process (MDP):
\begin{itemize}
    \item \textbf{State} $s_t = (q, a_{<t})$, representing the prompt and generated tokens.
    \item \textbf{Action} $a_t \in \mathcal{A}$, the token selected at step $t$.
    \item \textbf{Transition} $s_{t+1} = s_t \cup \{a_t\}$, deterministic concatenation.
\end{itemize}

Given a reward function $r(q, a)$ that evaluates the complete sequence (e.g., correctness or helpfulness), the standard RL objective is to maximize the expected reward:
\begin{equation}
J(\theta) = \mathbb{E}_{q \sim \mathcal{D},\, a \sim \pi_\theta(\cdot \mid q)} \big[ r(q, a) \big].
\label{eq:rl_objective}
\end{equation}

\paragraph{Entropy-regularized RL}
Standard RL methods often suffer from premature convergence to suboptimal deterministic policies. To mitigate this, the objective is typically augmented with an entropy regularization term.

We define the entropy of the policy distribution over the action space $\mathcal{A}$ at step $t$ as:
\begin{equation}
\mathcal{H}(\pi_\theta(\cdot \mid s_t))
:= - \sum_{a \in \mathcal{A}} \pi_\theta(a \mid s_t) \log \pi_\theta(a \mid s_t).
\label{eq:policy_entropy}
\end{equation}

To incorporate this into the loss minimization framework, we define the entropy regularization loss $\mathcal{L}_{\text{ent}, t}$ at step $t$ as the negative entropy of the policy:
\begin{equation}
\mathcal{L}_{\text{ent}, t}(\theta)
= - \mathcal{H}(\pi_\theta(\cdot \mid s_t)).
\label{eq:entropy_loss}
\end{equation}

The training objective at each step $t$ minimizes the combined loss:
\begin{equation}
\mathcal{L}_{\text{total}, t}(\theta) = \mathcal{L}_{\text{surr}, t}(\theta) + \alpha \mathcal{L}_{\text{ent}, t}(\theta),
\label{eq:total_loss}
\end{equation}
where $\mathcal{L}_{\text{surr}, t}(\theta)$ represents the step-level surrogate loss designed to maximize the expected reward $J(\theta)$ (e.g., PPO), and $\alpha$ is a hyperparameter controlling the strength of the regularization. This formulation encourages the policy to maintain stochasticity globally across the action space $\mathcal{A}$ at each generation step.

\section{Why Standard Entropy Regularization Fails in LLM RL}
\label{sec:method_failure}

While entropy regularization has demonstrated significant success in encouraging exploration for standard reinforcement learning benchmarks, such as continuous control or grid-world navigation, recent findings suggest it yields negligible effects or even degrades performance in the context of LLMs~\citep{cui2025entropy}. In this section, we analyze this phenomenon, attributing the failure to the interaction between the \textbf{sparsity of valid manifolds} and the \textbf{cumulative risk in long-horizon generation}.

\paragraph{The Curse of Dimensionality and Cumulative Risk}
The fundamental objective of entropy regularization is to prevent premature convergence by penalizing deterministic policies. While theoretically sound, its efficacy depends on the structural properties of the MDP. We highlight two critical disparities between classical RL and LLMs:

\textit{Sparse Manifolds and Tail Leakage.}
In classical domains (e.g., Atari), the action space is low-dimensional ($|\mathcal{A}| \approx 10$). A random walk often results in valid state transitions, allowing the agent to stumble upon rewards.
In contrast, LLMs operate over a massive vocabulary ($|\mathcal{A}| \approx 150,000$). The valid tokens that maintain syntactic coherence and logical continuity constitute an extremely sparse manifold.
Standard entropy regularization indiscriminately flattens the distribution. It shifts probability mass not just among plausible candidates, but explicitly leaks significant aggregate mass into the vast heavy tail of syntactically or logically invalid tokens.
While the probability increase for any single invalid token is negligible, the sum of probability mass over $10^5$ invalid candidates becomes substantial. This functionally injects high-variance noise into the decision process, rather than structured exploration.

\textit{The Survival Rate of Reasoning Chains.}
This leakage is exacerbated by the generation horizon $T$.
Let $\epsilon$ be the aggregate probability mass leaked to invalid tokens at each step due to entropy regularization. The probability of generating a coherent reasoning chain of length $T$ scales as $(1 - \epsilon)^T$.
In grid-worlds, a bad action is often recoverable. In LLM reasoning, a single semantic error in a Chain-of-Thought often renders the entire subsequent trajectory invalid.
For large $T$, even a minuscule $\epsilon$ guarantees that the policy will diverge from the valid reasoning manifold.

\paragraph{Impact of Generation Length: A Budget Constraint Analysis}
To empirically verify the correlation between horizon length and regularization failure, we conduct an analysis using Qwen2.5-1.5B-Instruct on the Maximum Generation Length ($T$) across MATH and Countdown tasks.
Here, $T$ represents the generation budget. The results in Figure~\ref{fig:entropy_length_comparison} reveal a crossover in efficacy governed by the interplay between noise intensity and sequence length.

\textit{High Error Tolerance Regime (Short Horizon).}
When the generation budget is constrained ($T=512$ for MATH; $T=32$ for Countdown), we observe that mild entropy regularization ($\alpha \in \{0.001, 0.0001\}$) consistently yields performance gains. 
In this regime, the trajectory is sufficiently short such that the cumulative probability of deviation remains low, allowing the model to benefit from local exploration. 
However, the fragility of this balance is evident in the Countdown task: overly aggressive regularization ($\alpha=0.01$) causes immediate degradation ($\Delta = -30.0\%$) even at the shortest horizon ($T=32$).

\begin{figure}[t]
    \centering
    \includegraphics[width=0.9\textwidth, height=6.2cm, keepaspectratio]{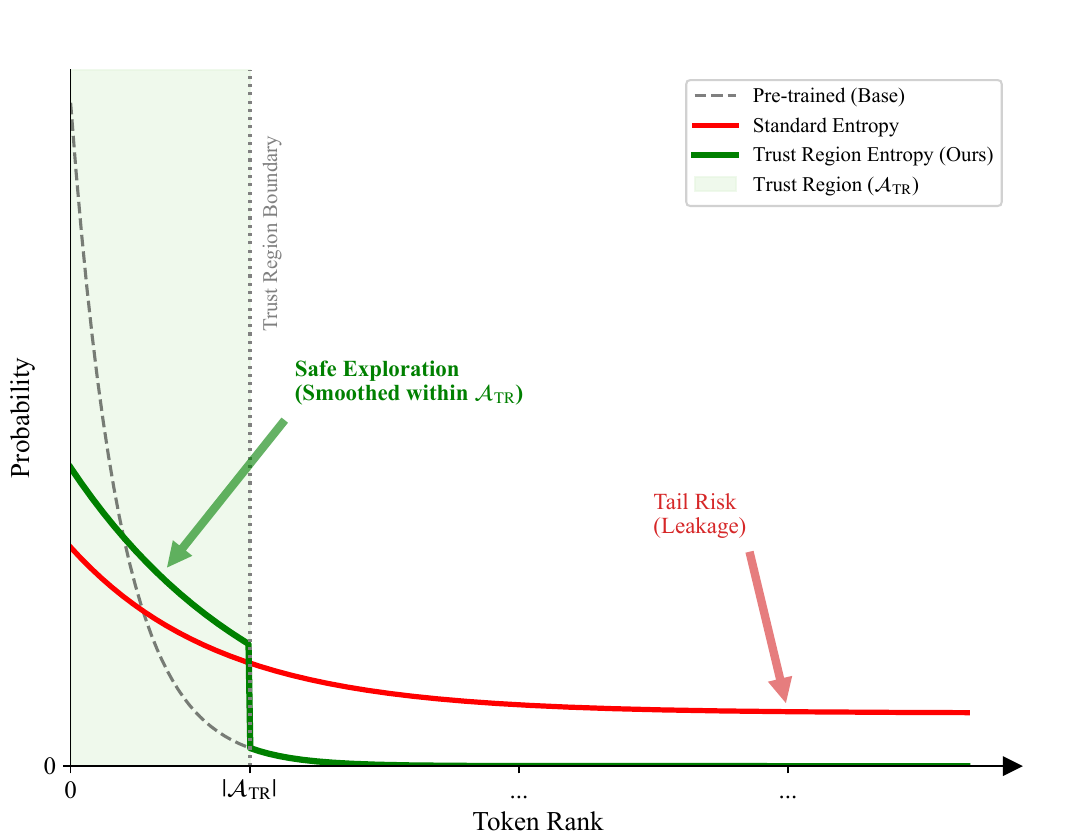}
    \vspace{-0.1cm} 
    \caption{\textbf{Overview of Trust Region Entropy (TRE).} Standard entropy regularization (red line) indiscriminately flattens the distribution, causing significant probability mass to leak into the vast tail of invalid tokens (Tail Risk). In contrast, TRE (green line) increases diversity exclusively within the Trust Region $\mathcal{A}_{\text{TR}}$ while maintaining the suppression of the Tail.}
    \label{fig:tre_diagram}
\end{figure}

\textit{Reasoning Collapse Regime (Long Horizon).}
As the generation budget extends to accommodate complex reasoning, a critical divergence emerges. The vanilla baseline (grey dashed line) effectively utilizes the extended context, showing a monotonic performance increase as $T$ grows (reaching $57.04\%$ on MATH and $64.12\%$ on Countdown). 
\textbf{However, entropy regularization fails to capitalize on this expanded capacity, effectively negating the benefits of the longer horizon}.
On the MATH task, while mild regularization was beneficial at short horizons, it becomes detrimental as the sequence lengthens, and at $T \ge 4096$, $\Delta$ turns negative for all $\alpha$ values, indicating that the cumulative risk accumulation outweighs any exploration benefit.
This collapse is even more pronounced in the Countdown task. At $T=512$, the cumulative risk of deviation leads to a catastrophic failure for the strong regularization setting ($\alpha=0.01$), resulting in a $\Delta$ of $-64.1\%$. Crucially, even the mildest regularization ($\alpha=0.0001$), which yielded significant gains at $T=32$, fails to maintain positive contribution at $T=512$. 
\textbf{This confirms that in long-context reasoning, the survival rate of a coherent chain under global entropy maximization drops precipitously, rendering valid reasoning statistically impossible}.


\section{Method}
\label{sec:method}

Standard entropy regularization indiscriminately promotes a uniform probability distribution across the entire vocabulary $\mathcal{A}$. In the context of LLMs, however, the vast majority of the token space comprises syntactically or semantically invalid options. As illustrated in Figure~\ref{fig:tre_diagram}(red curve), standard regularization forces the model to shift probability mass toward this tail, creating noise that degrades the stability and quality of the generation process.

To address this, we propose a targeted exploration mechanism inspired by the principles of \textit{Trust Region Policy Optimization} (TRPO). In TRPO, policy updates are strictly confined within a geometric region where the local approximation is trustworthy. Analogously, and consistent with the insight that truncation strategies isolate the reliable nucleus of a distribution \citep{Holtzman2020The}, we define a \textbf{Trust Region (TR)}, denoted as $\mathcal{A}_{\text{TR}}(s_t) \subset \mathcal{A}$. This region comprises only the most plausible candidate tokens identified by the model. We then encourage exploration exclusively within this trust region (see Figure~\ref{fig:tre_diagram}, green curve).

\paragraph{Defining the Trust Region}
The trust region $\mathcal{A}_{\text{TR}}(s_t)$ can be constructed using different filtering strategies. In this work, we consider two primary variants:

\begin{itemize}
    \item \textbf{TRE-K:} A \textbf{size-based} strategy where the trust region consists of the fixed $K$ tokens with the highest logit values at step $t$. This enforces a constant region size $|\mathcal{A}_{\text{TR}}(s_t)| = K$.
    \item \textbf{TRE-P:} A \textbf{nucleus-based} strategy where the trust region is the smallest set of tokens whose cumulative probability mass reaches a threshold $P \in (0, 1]$. Unlike \textit{TRE-K}, this defines a dynamic region size $|\mathcal{A}_{\text{TR}}(s_t)|$ that varies across time steps based on model confidence.
\end{itemize}

\paragraph{Trust Region Entropy (TRE)}
Given the full logits $\mathbf{z}_t$ and the identified trust region $\mathcal{A}_{\text{TR}}(s_t)$, we extract the corresponding sub-vector of logits $\tilde{\mathbf{z}}_t \in \mathbb{R}^{|\mathcal{A}_{\text{TR}}(s_t)|}$. To assess the diversity within this region, we compute a local probability distribution $\pi_{\text{local}}$ by applying the softmax function strictly over $\mathcal{A}_{\text{TR}}(s_t)$:

\begin{equation}
\pi_{\text{local}}(a \mid s_t) = \frac{\exp(z_{t,a})}{\sum_{a' \in \mathcal{A}_{\text{TR}}(s_t)} \exp(z_{t,a'})}, \quad \forall a \in \mathcal{A}_{\text{TR}}(s_t).
\label{eq:local_prob}
\end{equation}

We compute the entropy of this local distribution as:
\begin{equation}
\mathcal{H}(\pi_{\text{local}}(\cdot \mid s_t)) = - \sum_{a \in \mathcal{A}_{\text{TR}}(s_t)} \pi_{\text{local}}(a \mid s_t) \log \pi_{\text{local}}(a \mid s_t).
\end{equation}
Notably, if the trust region contains only a single token ($|\mathcal{A}_{\text{TR}}(s_t)| = 1$), which can occur in \textit{TRE-P} when the model is highly confident, the local entropy naturally becomes zero, effectively disabling regularization for that step.

To align the regularization magnitude with the full vocabulary entropy scale ($\log |\mathcal{A}|$), we introduce a scaling factor. For a trust region of size $|\mathcal{A}_{\text{TR}}(s_t)|$, the TRE loss at step $t$ is defined as:

\begin{equation}
\mathcal{L}_{\text{TRE}, t}(\theta) = 
\begin{cases} 
    - \frac{\log |\mathcal{A}|}{\log |\mathcal{A}_{\text{TR}}(s_t)|} \mathcal{H}(\pi_{\text{local}}) & \text{if } |\mathcal{A}_{\text{TR}}| > 1, \\[8pt]
    0 & \text{if } |\mathcal{A}_{\text{TR}}| = 1
\end{cases}
\label{eq:tre_loss}
\end{equation}

For \textit{TRE-K}, $|\mathcal{A}_{\text{TR}}(s_t)|$ is a constant $K$. For \textit{TRE-P}, $|\mathcal{A}_{\text{TR}}(s_t)|$ is the number of tokens in the nucleus. The final training objective at step $t$ is:
\begin{equation}
\mathcal{L}_{\text{total}, t}(\theta) = \mathcal{L}_{\text{surr}, t}(\theta) + \alpha \mathcal{L}_{\text{TRE}, t}(\theta).
\label{eq:total_loss_final}
\end{equation}

The complete computational procedure is summarized in Algorithm~\ref{alg:tre}.

\begin{algorithm}[t]
\caption{Trust Region Entropy (TRE)}
\label{alg:tre}
\begin{algorithmic}[1]
   \STATE {\bfseries Input:} Logits $\mathbf{z} \in \mathbb{R}^{|\mathcal{A}|}$, Strategy (TRE-K or TRE-P), Hyperparameter ($K$ or $P$)
   \STATE {\bfseries Output:} Loss $\mathcal{L}_{\text{TRE}, t}$
   \IF{Strategy is TRE-K}
   \STATE $\mathcal{I} \leftarrow \text{Indices of top } K \text{ values in } \mathbf{z}$
   \ELSIF{Strategy is TRE-P}
   \STATE $\mathbf{\pi} \leftarrow \text{Softmax}(\mathbf{z})$
   \STATE $\mathcal{I} \leftarrow \text{Smallest set of indices s.t. } \sum_{i \in \mathcal{I}} \pi_i \geq P$
   \ENDIF
   \IF{$|\mathcal{I}| \le 1$}
   \STATE \textbf{return} 0
   \ENDIF
   \STATE $\tilde{\mathbf{z}} \leftarrow \mathbf{z}[\mathcal{I}]$
   \STATE $\boldsymbol{\pi}_{\text{local}} \leftarrow \text{Softmax}(\tilde{\mathbf{z}})$  
   \STATE $\gamma \leftarrow \log |\mathcal{A}| \,/\, \log |\mathcal{I}|$  
   \STATE $\mathcal{L}_{\text{TRE}, t} \leftarrow \gamma \cdot \sum_{i=1}^{|\mathcal{I}|} -\boldsymbol{\pi}_{\text{local}, i} \log \boldsymbol{\pi}_{\text{local}, i}$ 
   \STATE \textbf{return} $\mathcal{L}_{\text{TRE}, t}$
\end{algorithmic}
\end{algorithm}


\begin{table*}[t]
    \centering
    \caption{\textbf{Main Results.} Comparison of different methods on Qwen2.5-1.5B-Instruct and Qwen2.5-7B-Instruct. \textbf{Base} refers to the base model before reinforcement learning. \textbf{Vanilla} denotes the standard PPO baseline. Values in parentheses indicate the relative performance change ($\Delta$) compared to the \textbf{Vanilla} baseline. \textbf{Bold} and \underline{underline} indicate the best and second-best performance, respectively. Note that the Base model is provided for context and is not included in the ranking or $\Delta$ calculation.}
    \label{tab:main_results}
    \vspace{-0.1cm}
    \setlength{\tabcolsep}{5pt}
    \begin{tabular}{l c c c c c c}
        \toprule
        \multirow{3}{*}{\textbf{Method}} 
        & \multicolumn{3}{c}{\textbf{Qwen2.5-1.5B-Instruct}} 
        & \multicolumn{3}{c}{\textbf{Qwen2.5-7B-Instruct}} \\
        \cmidrule(lr){2-4} \cmidrule(lr){5-7}
        & \textbf{MATH} & \textbf{Countdown} & \textbf{HH} 
        & \textbf{MATH} & \textbf{Countdown} & \textbf{HH} \\
        & \small{\textit{Pass@1 (\%)}} & \small{\textit{Pass@1 (\%)}} & \small{\textit{Reward}} 
        & \small{\textit{Pass@1 (\%)}} & \small{\textit{Pass@1 (\%)}} & \small{\textit{Reward}} \\
        \midrule
        \textbf{Base} 
            & 51.06& 2.54& 0.0036& 71.42 & 31.66 & 2.0749 \\ 
        \midrule
        \textbf{Vanilla (PPO)}   
            & 57.04 & 64.12 & 3.24 & 73.52 & 72.38 & 2.8839 \\ 
        \midrule
        Ent   
            & 56.64 \loss{-0.4}& 63.20 \loss{-0.92}& 3.1913 \loss{-0.0487} 
            & 73.66 \gain{0.14} & 72.60 \gain{0.22} & 2.8718 \loss{-0.0121} \\
        Forking-Tokens         
            & 57.16 \gain{0.12} & 62.82 \loss{-1.3} & 3.3184 \gain{0.0784} 
            & 73.59 \gain{0.07} & 72.94 \gain{0.56} & 2.8490 \loss{-0.0349} \\
        KL-Cov        
            & 58.23 \gain{1.19} & \underline{66.5} \gain{2.38} & 3.3931 \gain{0.1531} 
            & 73.79 \gain{0.27} & 73.4 \gain{1.02} & 2.9529 \gain{0.069} \\
        \midrule
        \textbf{TRE-K (Ours)} 
            & \underline{58.26} \gain{1.22} & 66.28 \gain{2.16} & \underline{3.8209} \gain{0.5809} 
            & \textbf{74.29} \gain{0.77} & \underline{74.76} \gain{2.38} & \underline{2.9715} \gain{0.0876} \\
        \textbf{TRE-P (Ours)} 
            & \textbf{58.28} \gain{1.24} & \textbf{66.96} \gain{2.84} & \textbf{3.8764} \gain{0.6364} 
            & \underline{74.27} \gain{0.75} & \textbf{75.3} \gain{2.92} & \textbf{3.0331} \gain{0.1492} \\
        \bottomrule
    \end{tabular}
\end{table*}

\section{Experimental Setup}
\label{sec:setup}

\paragraph{Models and Implementation Details}
We conduct our experiments using the Qwen2.5-1.5B-Instruct and Qwen2.5-7B-Instruct as the base language models~\cite{qwen2.5}. For the post-training alignment phase, we employ PPO algorithm, implemented within the VeRL framework~\cite{sheng2024hybridflow}. The training is performed with a global batch size of 512. During the reinforcement learning sampling process, we set the temperature to 1.0 and the top-$p$ to 1.0. For our proposed TRE, we evaluate two variants: \textit{TRE-K} with a fixed trust region size of $K=2$, and \textit{TRE-P} with a cumulative probability threshold of $P=0.99$. We set the regularization coefficient at $\alpha=0.001$ for both variants across all tasks. Further implementation details and hyperparameter settings are provided in Appendix~\ref{detail_experiments}.

\paragraph{Tasks and Datasets}
To evaluate the model's performance across diverse capabilities, we conduct experiments on three distinct tasks: mathematical reasoning, combinatorial search, and general preference alignment. We curate specific training and testing splits for each task as follows:

\begin{itemize}
    \item \textbf{MATH Task:} We utilize the MATH dataset~\cite{hendrycksmath2021}, which compiles problems from various mathematics competitions (e.g., AMC, AIME). The dataset covers seven subject areas: Algebra, Intermediate Algebra, Prealgebra, Geometry, Number Theory, Counting \& Probability, and Precalculus. It consists of 7,500 examples for training and 5,000 examples for testing. For this task, the maximum generation length is set to 8,192 tokens.
    
    \item \textbf{Countdown Task:} The objective is to generate mathematical equations using $N$ given numbers to reach a specific target value. We source the data from the Countdown Tasks dataset~\cite{tinyzero}. From this source, we randomly sample 7,500 instances to construct the training set and 5,000 instances for the test set. The maximum generation length for this task is limited to 512 tokens.
    
    \item \textbf{HH (Helpful and Harmless) Task:} For general instruction following and human preference alignment, we employ the UltraFeedback dataset~\cite{cui2023ultrafeedback}. From this source, we randomly select 7,500 samples for the training set and 5,000 samples for the test set to evaluate the model's alignment performance. The maximum generation length is set to 1,024 tokens.
\end{itemize}

\paragraph{Reward and Evaluation}
We maintain strict consistency between the judging mechanisms used during training and evaluation. For \textbf{MATH} and \textbf{Countdown}, we employ a rule-based evaluator that checks the output format and extracts the final answer. We assign a binary reward of 1 for a correct answer and 0 for an incorrect one. For \textbf{HH}, we utilize the scalar score provided by a learned reward model. We trained separate reward models on the full dataset for each backbone model, and specific training details are provided in Appendix~\ref{detail_experiments}.

To obtain reliable performance estimates, we independently sample 8 responses per prompt for all test sets, utilizing a temperature of 1.0 and a top-$p$ of 0.7. We report the Pass@1 metric for the MATH and Countdown tasks, while the average reward score for the HH tasks. It is worth noting that our experiments focus on analyzing the performance of reinforcement learning algorithms within a controlled environment. Consequently, we treat the reward signals as the ground-truth objective and consider the analysis of reward hacking to be beyond the scope of this work.

\paragraph{Comparison}
\begin{itemize}
    \item \textbf{Vanilla / Ent}: These variants follow the entropy-regularized objective defined in Eq.~\ref{eq:total_loss}. \textbf{Vanilla} represents the standard PPO algorithm without entropy regularization ($\alpha=0$). \textbf{Ent} incorporates entropy regularization to encourage exploration and prevent premature convergence; we set the coefficient $\alpha=0.001$, as evidence in Section~\ref{sec:method_failure} and prior work~\cite{cui2025entropy} demonstrates this value yields optimal performance.

    \item \textbf{Forking-Tokens}~\cite{wang2025beyond}: concentrates optimization strictly on \textit{Forking Tokens} by applying policy updates only to the top $\gamma=20\%$ of steps with the highest entropy within a batch. The step-level loss is defined as: $\mathcal{L}_{\text{total}, t}(\theta) = \mathbb{I}(\mathcal{H}(\pi_\theta(\cdot \mid s_t)) \in \text{Top}_{\gamma}) \cdot \mathcal{L}_{\text{surr}, t}(\theta)$.

    \item \textbf{KL-Cov}~\cite{cui2025entropy}: selectively imposes a strong KL penalty ($\beta=1.0$) solely on the top $\gamma=20\%$ of tokens that exhibit the highest covariance between advantage estimates and log-probabilities. The step-level loss is defined as: $\mathcal{L}_{\text{total}, t}(\theta) = \mathcal{L}_{\text{surr}, t}(\theta) + \mathbb{I}(t \in \text{Top}_{\gamma}(\text{Cov}(A, \log \pi))) \cdot \beta D_{\text{KL}}(\pi_{\text{old}} \parallel \pi_\theta)_t$.

\end{itemize}


\begin{figure*}[t]
    \centering
    \begin{subfigure}[b]{0.32\textwidth}
        \centering
        \includegraphics[width=\textwidth]{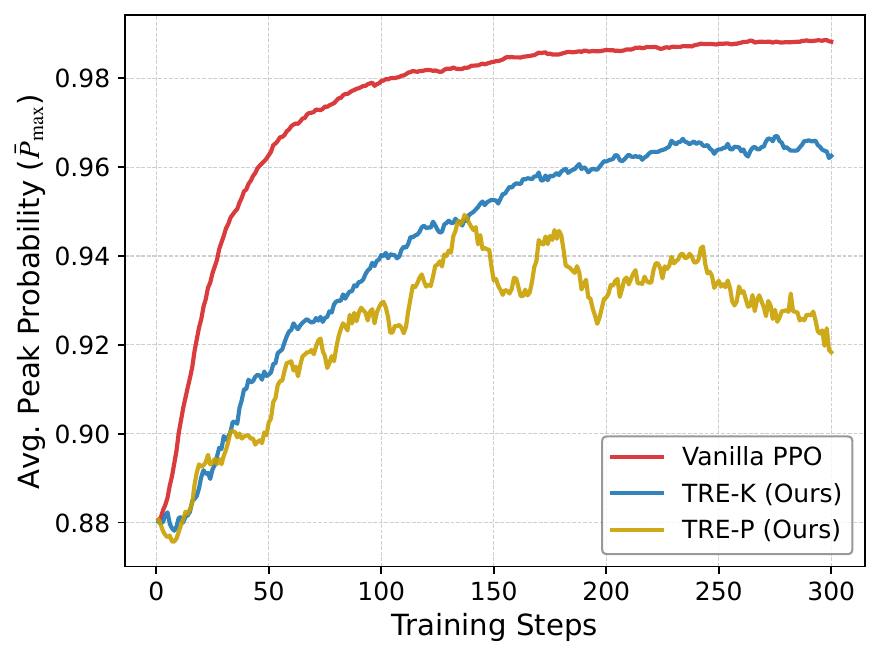} 
        \centerline{\small \textbf{(a) MATH Task}}
    \end{subfigure}
    \hfill
    \begin{subfigure}[b]{0.32\textwidth}
        \centering
        \includegraphics[width=\textwidth]{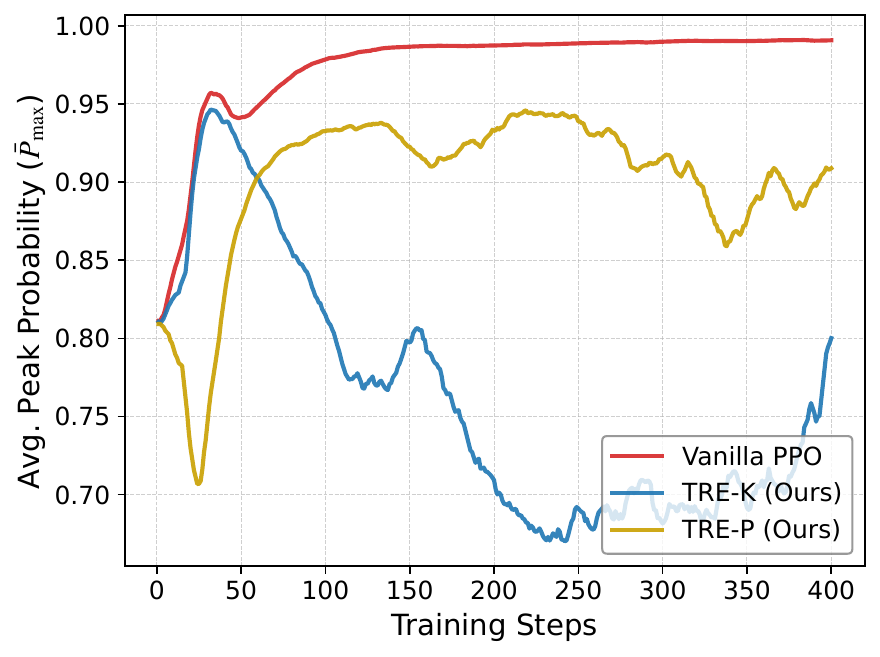}
        \centerline{\small \textbf{(b) Countdown Task}}
    \end{subfigure}
    \hfill
    \begin{subfigure}[b]{0.32\textwidth}
        \centering
        \includegraphics[width=\textwidth]{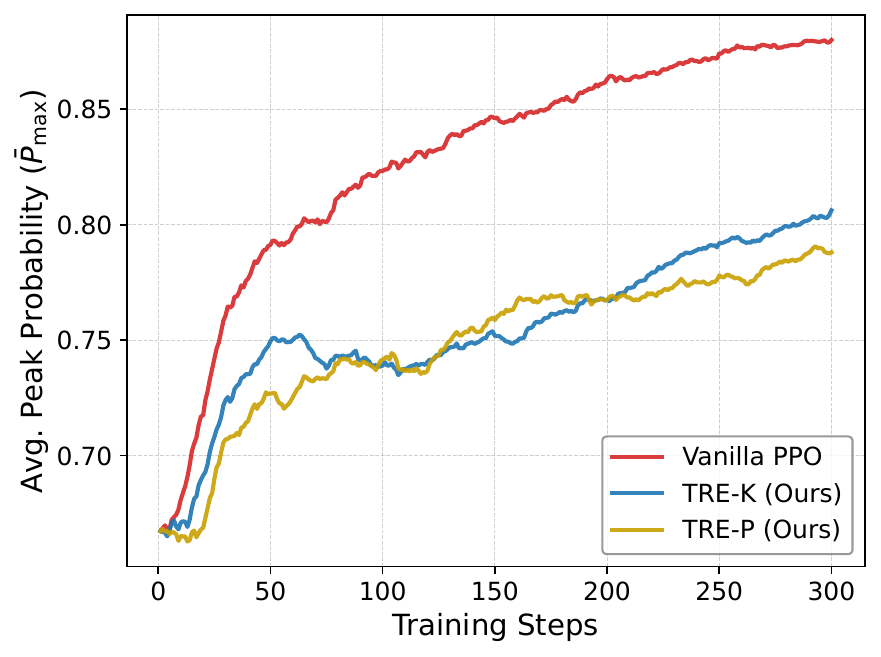}
        \centerline{\small \textbf{(c) HH Task}}
    \end{subfigure}
    \vspace{-0.1cm}
    \caption{\textbf{Evolution of Average Peak Probability ($\bar{P}_{\max}$) on Qwen2.5-1.5B-Instruct.} The curves illustrate the policy's confidence level during training across three tasks. The \textbf{Vanilla PPO} baseline (Red) consistently trends toward saturation (high confidence), while \textbf{TRE-K} (Blue) and \textbf{TRE-P} (Gold) maintain a lower peak probability. A smoothing factor of 0.8 is applied for visualization clarity.}
    \label{fig:confidence_dynamics}
\end{figure*}

\section{Results}
\label{result}

\subsection{Main Results}
\label{sec:main_result}

Table~\ref{tab:main_results} presents the primary performance comparison across three diverse tasks. Our proposed Trust Region Entropy (TRE) methods, specifically \textit{TRE-K} and \textit{TRE-P}, consistently outperform the \textit{Vanilla} baseline and existing regularization techniques across both the 1.5B and 7B model scales.

\paragraph{Limitations of Ent and Forking-Tokens} 
As shown in Table~\ref{tab:main_results}, both standard entropy regularization (\textit{Ent}) and the selective update scheme \textit{Forking-Tokens} exhibit limited and inconsistent gains across tasks and model scales. 

\textit{Ent} shows mixed behavior. While it brings slight improvements in some settings, such as gains of 0.14\% on 7B MATH and 0.22\% on 7B Countdown, it degrades performance in four out of six cases. These failures include a 0.92\% drop on 1.5B Countdown and a reduction of 0.0487 in the 1.5B HH reward. This results indicates that global entropy regularization introduces significant tail risk that often offsets or even outweighs the benefits of exploration, particularly in long-horizon reasoning where cumulative errors are amplified.

In contrast, \textit{Forking-Tokens} does not introduce an explicit entropy loss. It instead restricts optimization to the top-$\gamma$ high-entropy steps and exempts low-entropy steps from surrogate loss. Although this selective surrogate-loss strategy yields marginal gains in some scenarios, it also underperforms the \textit{Vanilla} baseline in others, such as a 1.3\% regression on 1.5B Countdown and a 0.0349 decrease in the 7B HH reward. These results indicate that maintaining exploratory behavior solely through surrogate loss by exempting low-entropy steps cannot stably achieve performance improvements.

\paragraph{TRE vs. the Strongest Baseline (KL-Cov)} 
\textit{KL-Cov} serves as the most competitive baseline, particularly on reasoning tasks at the 1.5B scale. In the 1.5B MATH task, \textit{KL-Cov} achieves a gain of +1.19\%, which is closely comparable to our +1.24\%. Notably, on the 1.5B Countdown task, \textit{KL-Cov} (+2.38\%) achieves the second-best performance, even surpassing our \textit{TRE-K} variant (+2.16\%). 

However, the superiority of our TRE framework becomes markedly more pronounced in the alignment task and at larger model scales. Specifically, in the 1.5B HH task, \textit{TRE-P} delivers a reward gain approximately \textbf{4.1 times} larger than that of \textit{KL-Cov} (+0.6364 vs. +0.1531). This trend continues across all 7B tasks, where \textit{TRE-P} consistently produces improvements over the \textit{Vanilla} baseline that are \textbf{2 to 3 times} greater than those of \textit{KL-Cov} (e.g., +2.92\% vs. +1.02\% on Countdown, and +0.1492 vs. +0.069 on HH). This demonstrates that by explicitly regularizing the trust region, TRE provides a much more robust and scalable exploration signal than methods relying on covariance-based step selection.

\paragraph{Comparison between TRE-K and TRE-P} 
Among the two proposed variants, \textit{TRE-P} emerges as the overall top-performing method, achieving the best results in five out of six evaluation categories. By dynamically adjusting the trust region size based on cumulative probability, \textit{TRE-P} adapts more flexibly to varying levels of model confidence. In the 7B MATH task, while \textit{TRE-K} (74.29\%) marginally outperforms \textit{TRE-P} (74.27\%), the difference is a negligible 0.02\%. The overall stability of \textit{TRE-P} across tasks—especially its significant leads in the HH reward—positions it as the most effective strategy for balancing diversity and precision.

\begin{figure*}[htbp]
    \centering
    \begin{subfigure}[b]{0.32\textwidth}
        \centering
        \includegraphics[width=\textwidth]{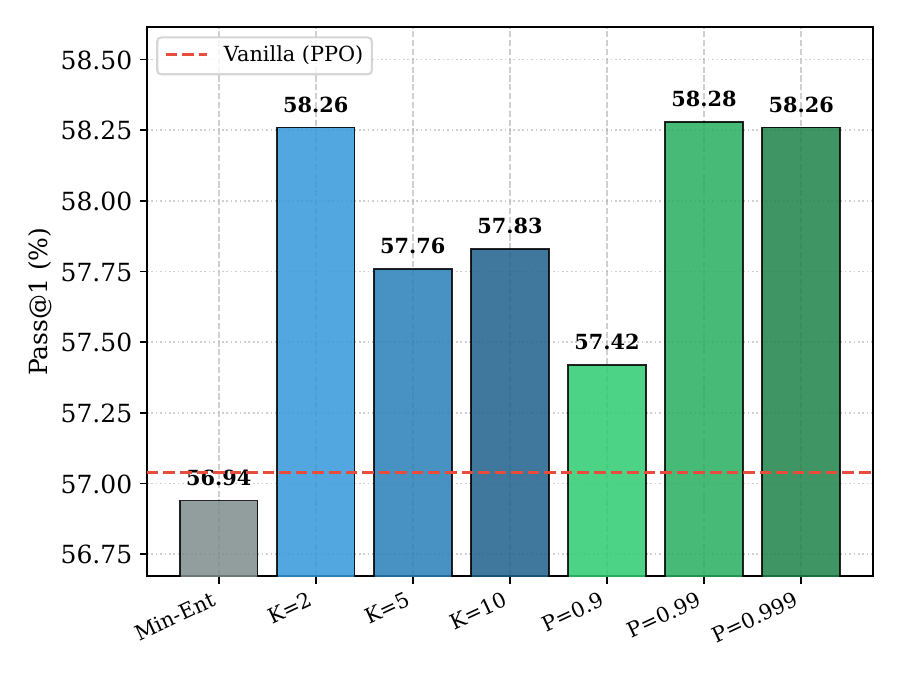}
        \centerline{\small \textbf{(a) MATH Task}}
        \label{fig:math_ablation}
    \end{subfigure}
    \hfill
    \begin{subfigure}[b]{0.32\textwidth}
        \centering
        \includegraphics[width=\textwidth]{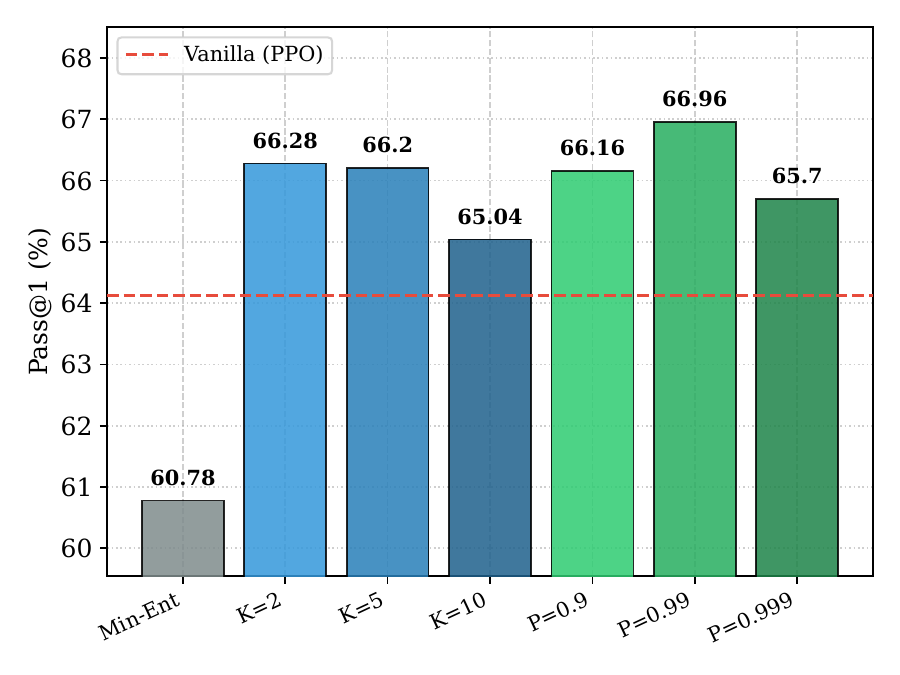}
        \centerline{\small \textbf{(b) Countdown Task}}
        \label{fig:countdown_ablation}
    \end{subfigure}
    \hfill
    \begin{subfigure}[b]{0.32\textwidth}
        \centering
        \includegraphics[width=\textwidth]{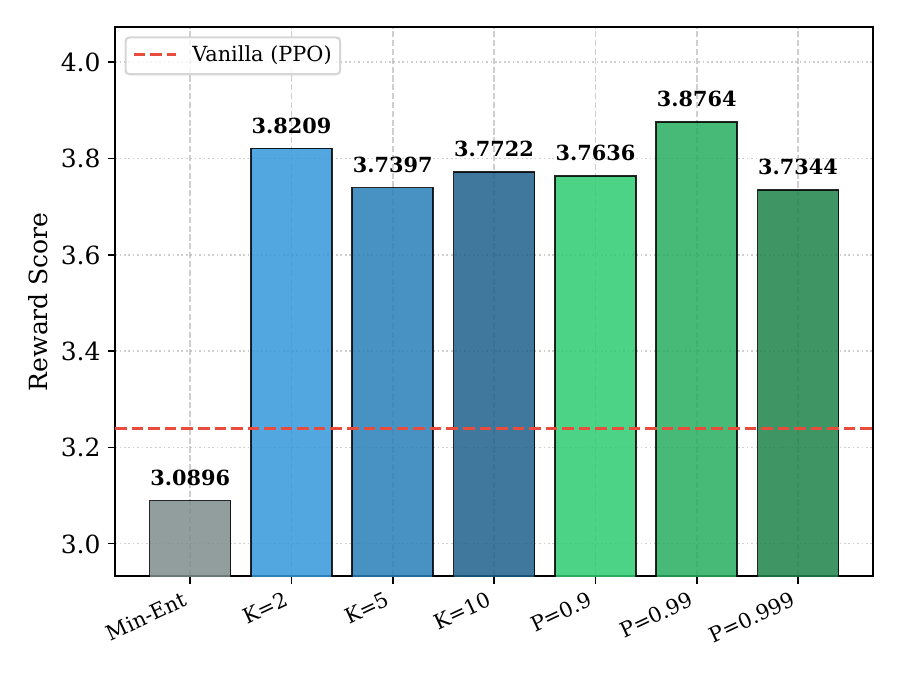}
        \centerline{\small \textbf{(c) HH Task}}
        \label{fig:hh_ablation}
    \end{subfigure}
    \caption{\textbf{Hyperparameter Analysis on Qwen2.5-1.5B-Instruct.} We compare the performance of size-based trust regions ($K$), nucleus-based trust regions ($P$), and the Min-Ent baseline ($K=1$ boundary). The dashed red line represents the \textbf{Vanilla (PPO)} performance without entropy regularization.}
    \label{fig:ablation_all}
\end{figure*}

\subsection{Policy Confidence Dynamics}
\label{sec:confidence_analysis}

To more accurately diagnose the issue of premature convergence during RL training, we introduce the \textbf{Average Peak Probability} ($\bar{P}_{\max}$). This metric tracks the model's confidence in its top-1 prediction. We define $\bar{P}_{\max}$ as the expected maximum probability over the states $s_t$ within a training batch:
\begin{equation}
\bar{P}_{\max} = \mathbb{E}_{s_t} \left[ \max_{a \in \mathcal{A}} \pi_\theta(a \mid s_t) \right].
\label{eq:peak_prob}
\end{equation}
$\bar{P}_{\max}$ approaching $1.0$ indicates that the policy has become nearly deterministic, effectively ceasing exploration.


Figure~\ref{fig:confidence_dynamics} illustrates the evolution of $\bar{P}_{\max}$ across the MATH, Countdown, and HH tasks. A distinct pattern emerges: the \textit{Vanilla} baseline (Red line) exhibits a rapid and aggressive increase in peak probability. Particularly in the MATH and Countdown tasks (Figures~\ref{fig:confidence_dynamics}a and b), the PPO curve quickly saturates near 0.99. This trajectory indicates that the standard RL objective forces the model to become overconfident early in the optimization process, prematurely locking into a narrow reasoning path. 

We attribute this aggressive saturation to the \textbf{binary nature of the reward signals} in reasoning tasks, where the strict dichotomy between correct and incorrect answers encourages the model to collapse its distribution onto a single valid path. In contrast, the HH task (Figure~\ref{fig:confidence_dynamics}c), driven by a continuous preference model, exhibits a slower and lower rise in confidence ($\sim$0.88). This reflects the inherent ambiguity and nuance of human alignment, which offers a smoother optimization landscape compared to the rigid verifiability of mathematical problems.

In contrast, our proposed methods effectively mitigate this confidence collapse, though with notable differences between the variants:

\begin{itemize}
    \item \textbf{MATH Task} (Fig.~\ref{fig:confidence_dynamics}a): Both TRE methods prevent the model from converging to a deterministic state. Notably, \textit{TRE-P} (Gold) maintains a lower and more diverse confidence level ($\sim$0.92) compared to \textit{TRE-K} (Blue, $\sim$0.96). This suggests that the dynamic nucleus approach encourages broader exploration than the fixed top-$K$ constraint, aligning with TRE-P's superior performance in Table~\ref{tab:main_results}.
    
    \item \textbf{Countdown Task} (Fig.~\ref{fig:confidence_dynamics}b): This task reveals a critical stability difference. While Vanilla PPO saturates at $\sim$0.99, \textit{TRE-K} exhibits significant instability, with confidence dropping drastically to $\sim$0.67 in the mid-training phase. This implies that forcing a fixed number of tokens can disrupt the model's policy when the ground-truth distribution is sharp. Conversely, \textit{TRE-P} demonstrates greater resilience, despite an initial fluctuation, it rapidly recovers and stabilizes at a healthy confidence level ($\sim$0.9), effectively avoiding the sustained collapse observed in \textit{TRE-K}.
    
    \item \textbf{HH Task} (Fig.~\ref{fig:confidence_dynamics}c): While Vanilla PPO climbs steadily to $\sim$0.88, both TRE variants suppress this overconfidence at $\sim$0.8. This larger gap in stochasticity is crucial for alignment tasks, as it allows the model to continuously explore subtle nuances in human preferences rather than greedily converging to a single response style.
\end{itemize}

These results confirm that TRE effectively mitigates policy overconfidence. Furthermore, the comparison in the Countdown task highlights the structural advantage of \textit{TRE-P}: by dynamically adapting the trust region size, it avoids the instability pitfalls of fixed-size regularization while preventing the deterministic saturation of standard PPO.

\subsection{Hyperparameter Analysis}
\label{sec:ablation_study}

To better understand the impact of the trust region's scope on model performance, we conduct a hyperparameter analysis on the Qwen2.5-1.5B-Instruct model. We vary the fixed trust region size $K \in \{2, 5, 10\}$ for \textit{TRE-K} and the cumulative probability threshold $P \in \{0.9, 0.99, 0.999\}$ for \textit{TRE-P}. 

\paragraph{Min-Ent} 
Since \textit{TRE-K} achieves optimal performance at $K=2$, we further investigate the boundary condition where exploration is constrained to the top-1 prediction ($K=1$). To handle this case, we employ \textit{Min-Ent} utilizing Min-entropy $\mathcal{H}_{\infty}(\pi) = -\log (\max_{a \in \mathcal{A}} \pi(a|s_t))$, which is a special case of the Rényi entropy. The total objective is defined as $\mathcal{L}_{\text{total}, t}(\theta) = \mathcal{L}_{\text{surr}, t}(\theta) - \alpha \cdot \mathcal{H}_{\infty}(\pi_{\theta}(\cdot \mid s_t))$, where we maintain $\alpha=0.001$.


As illustrated in Figure~\ref{fig:ablation_all}, the choice of the trust region scope significantly influences the final performance. Our key observations are as follows:

First, \textbf{Min-Ent} consistently underperforms the Vanilla baseline across all tasks, achieving 56.94\% on MATH, 60.78\% on Countdown, and 3.0896 on HH. This performance degradation suggests that the Min-Ent regularization, while intended to encourage exploration, may excessively hinder exploitation by preventing the policy from confidently committing to its current best action, thereby hurting overall learning efficacy.

Second, for the size-based variant \textbf{TRE-K}, we observe that \textbf{$K=2$} consistently achieves the peak performance across all three tasks. Although a strict monotonic decline is most evident in the Countdown task (where Pass@1 drops from 66.28\% at $K=2$ to 65.04\% at $K=10$), a similar pattern of performance degradation relative to the $K=2$ peak is observed in MATH and HH. In these cases, while there are slight fluctuations as $K$ increases from 5 to 10, the model never regains the performance level achieved at $K=2$. This demonstrates that while moderate exploration among the top candidates is beneficial, expanding the trust region too far generally re-introduces low-probability noise, which dilutes the gradient signal and hampers overall model performance.

Third, for the nucleus-based variant \textbf{TRE-P}, the cumulative threshold \textbf{$P=0.99$} emerges as the optimal configuration, consistently outperforming both the \textit{Vanilla} baseline and all \textit{TRE-K} variants. Specifically, $P=0.9$ is too narrow, often performing worse than $K=2$. Conversely, $P=0.999$ begins to include invalid tail tokens that reduce scores, as seen in the drop from 66.96\% to 65.7\% in the Countdown task. Therefore, $P=0.99$ strikes the best balance. The fact that \textit{TRE-P} ($P=0.99$) achieves the overall highest scores, validates the advantage of a dynamic trust region: it allows the model to explore more broadly when it is uncertain and focus more tightly when it is confident, effectively maximizing exploration efficiency within a valid space.
\section{Discussion}
\label{sec:discussion}

Standard entropy regularization fails in LLMs due to the \textbf{cumulative tail risk} introduced by the vast vocabulary and long generation horizons. While it encourages exploration, it indiscriminately spreads probability mass across the entire vocabulary, including syntactically or semantically invalid tokens. This results in increased noise, which accumulates over long generation steps and disrupts coherent reasoning. In contrast, our \textbf{Trust Region Entropy (TRE)} method focuses exploration within a more manageable trust region of valid tokens, avoiding this issue and enhancing performance.

\paragraph{Limitations} Despite these promising results, several areas warrant further investigation. First, the efficacy of TRE over \textbf{ultra-long generation horizons} (e.g., 128k+ tokens) and its impact on long-range coherence in extensive CoT (Chain-of-Thought) trajectories remain to be verified. Second, while TRE scales effectively up to 7B parameters, its behavior and hyperparameter stability on \textbf{massive models} require further empirical study.


\bibliography{example_paper}
\bibliographystyle{icml2026}

\newpage
\appendix
\onecolumn
\section{Appendix}
\subsection{Implementation details of Experiments}
\label{detail_experiments}

\paragraph{Hyper-parameters for Training} We report the detailed hyperparameters used for PPO training in Table \ref{app_tab:training_hyperparams}. The actor and critic models use learning rates of $1\times10^{-6}$ and $1\times10^{-5}$, respectively. We set the global batch size to 512 and generate 8 rollouts per prompt. For Generalized Advantage Estimation (GAE), we set both $\gamma$ and $\lambda$ to 1.0.To ensure sufficient training, we run the optimization for 300 steps on the MATH task, 400 steps on the Countdown task, and 300 steps on the HH task.
Regarding the generation constraints, the maximum generation length for the HH task is capped at 1,024 tokens. This choice is specifically aligned with the sequence length distribution of the training data used for our reward models, ensuring that the policy operates within the well-calibrated domain of the scalar reward functions.

\begin{table}[h]
  \centering  
  \begin{tabular}{l c}
    \toprule
    \textbf{Parameter}      & \textbf{Value}    \\
    \midrule
    Actor Learning rate     & $1\times10^{-6}$  \\
    Critic Learning rate    & $1\times10^{-5}$  \\
    Global Batch size       & $512$              \\
    PPO Mini-batch size     & $64$               \\
    Number of Rollouts ($N$)& $8$                \\
    Clip Range              & $0.2$             \\
    KL Coefficient          & $0.0$            \\
    Loss Aggregation Mode   & token-mean       \\
    GAE Gamma ($\gamma$)    & $1.0$            \\
    GAE Lambda ($\lambda$)  & $1.0$            \\
    \bottomrule
  \end{tabular}
  \caption{Training hyperparameters for the PPO experiments.}
  \label{app_tab:training_hyperparams}
\end{table}

\paragraph{Evaluation and Reward Implementation} 
For the MATH task, we utilize the tools provided by \cite{eval-harness} for judgment, which involves extracting the answer within \texttt{\textbackslash boxed\{\}} for rule-based matching. For the Countdown task, we employ the tools from \cite{tinyzero} for evaluation, extracting the content within \texttt{\textless answer\textgreater\textless /answer\textgreater} tags to compute and verify the results. 

For the HH task, we employ the TRL framework~\cite{vonwerra2022trl} to train scalar reward models on the complete UltraFeedback dataset. To ensure consistent evaluation across different model scales, we trained two separate reward models based on Qwen2.5-1.5B-Instruct and Qwen2.5-7B-Instruct, respectively. Both models were trained for 3 epochs with a learning rate of $1\times10^{-5}$, a cosine scheduler with a 0.1 warmup ratio, and a center rewards coefficient of $1\times10^{-2}$. The global batch size was set to 512.

These reward models were calibrated to have an average reward score of approximately 0 on the training set. The 1.5B reward model achieves an accuracy of 96.64\% with scores distributed in the interval $[-3.5, 4.0]$, while the 7B reward model achieves an accuracy of 98.5\% with scores distributed in the interval $[-3.3, 3.3]$.

MATH task prompt template:
\begin{tcolorbox}[colback=gray!5!white,colframe=gray!75!black]
\textbf{User:} \{problem\} Let's think step by step and output the final answer within \texttt{\textbackslash boxed\{\}}.
\end{tcolorbox}
Countdown task prompt template:
\begin{tcolorbox}[colback=gray!5!white,colframe=gray!75!black]
\textbf{User:} Using the numbers \{numbers\}, create an equation that equals \{target\}. You can use basic arithmetic operations (+, -, *, /) and each number can only be used once. Show your work in \textless think\textgreater{} \textless /think\textgreater{} tags. And return the final answer in \textless answer\textgreater{} \textless /answer\textgreater{} tags, for example \textless answer\textgreater{} (1 + 2) / 3 \textless /answer\textgreater.
\end{tcolorbox}
HH task prompt template:
\begin{tcolorbox}[colback=gray!5!white,colframe=gray!75!black]
\textbf{User:} \{problem\}
\end{tcolorbox}

\paragraph{Experimental Scope and Constraints for HH Task}
For the HH task, the maximum generation length is fixed at 1,024 tokens, aligning with the maximum sequence length used during the training of the reward models to ensure the reliability of the reward signals. 
Furthermore, unlike reasoning-intensive tasks (MATH and Countdown) where the final output is a concise entity (e.g., a boxed value or a single equation) preceded by a long chain-of-thought, the HH task focuses on generating human-aligned, holistic responses where the entire sequence constitutes the answer. Given that the quality of these responses is intrinsically evaluated by a scalar reward model—which may exhibit distribution shift or saturation beyond its training horizon—we do not include the HH task in the \textit{Impact of Maximum Generation Length} analysis. This ablation is reserved for the reasoning domains where the performance can be objectively verified through rule-based matching across varying generation budgets.
\end{document}